\documentclass[runningheads]{llncs}

\usepackage[T1]{fontenc}
\usepackage{multirow}
\usepackage{caption} 
\DeclareCaptionLabelFormat{boldnumber}{\textbf{#1 #2}}
\usepackage{booktabs}
\usepackage[table,xcdraw]{xcolor}
\usepackage{pifont}
\usepackage{amsmath}
\usepackage{wasysym}
\usepackage{booktabs}
\usepackage{amssymb}
\usepackage{lmodern} % 使用现代字体
\usepackage{marvosym}
\usepackage{microtype}
\usepackage{graphicx}
\begin{document}

\title{BIMCV-R: A Landmark Dataset for 3D CT Text-Image Retrieval}
% \author{Paper ID: 499}
\author{Yinda Chen\inst{1,2} \and Che Liu\inst{3} \and Xiaoyu Liu\inst{1} \and Rossella Arcucci\inst{3} \and Zhiwei Xiong\inst{1,2}\textsuperscript{(\Letter)}}
\authorrunning{Chen et al.}

\institute{
    $^1$MoE Key Laboratory of Brain-inspired Intelligent Perception and Cognition, University of Science and Technology of China\\
    $^2$Anhui Province Key Laboratory of Biomedical Imaging and Intelligent Processing, Institute of Artificial Intelligence, Hefei Comprehensive National Science Center\\
    $^3$Data Science Institute, Imperial College London\\
    \email{cyd0806@mail.ustc.edu.cn},
    \email{zwxiong@ustc.edu.cn}
}

\maketitle              % typeset the header of the contribution
\begin{abstract}
The burgeoning integration of 3D medical imaging into healthcare has led to a substantial increase in the workload of medical professionals. To assist clinicians in their diagnostic processes and alleviate their workload, the development of a robust system for retrieving similar case studies presents a viable solution.
While the concept holds great promise, the field of 3D medical text-image retrieval is currently limited by the absence of robust evaluation benchmarks and curated datasets. To remedy this, our study presents a groundbreaking dataset, {BIMCV-R}, which includes an extensive collection of 8,069 3D CT volumes, encompassing over 2 million slices, paired with their respective radiological reports.
Expanding upon the foundational work of our dataset, we craft a retrieval strategy, MedFinder. This approach employs a dual-stream network architecture, harnessing the potential of large language models to advance the field of medical image retrieval beyond existing text-image retrieval solutions. It marks our preliminary step towards developing a system capable of facilitating text-to-image, image-to-text, and keyword-based retrieval tasks. Our project is available at \url{https://huggingface.co/datasets/cyd0806/BIMCV-R}.
% We hold the cautious optimism that our efforts may lay the groundwork for additional scholarly exploration and the refinement of techniques within the realm of 3D medical text-image retrieval.

\end{abstract}

\section{Introduction}
The rapid evolution of medical imaging technologies, especially in the realm of 3D imaging, has brought about a transformative change in radiological diagnostics \cite{yu2024high,abdulla2020meta,lai2024adaptive,li2024gtp}. These technologies provide detailed, three-dimensional visualizations that are crucial for accurate lesion detection and measurement, which are pivotal in disease staging, treatment planning, and prognosis assessment \cite{cheng2022characterization,smalheiser2021effect}. However, the sheer volume and complexity of this spatial data have substantially increased the workload of clinicians, prompting the need for Artificial Intelligence (AI) to assist in the analysis process. AI applications such as automatic segmentation \cite{hatamizadeh2021swin,chen2024tokenunify}, reconstruction \cite{sun2024data,chen2023self,yang2024unicompress,li2023steganerf}, and denoising \cite{deng2022unified} are now integral in providing clearer insights and improving diagnostic accuracy.

AI's role in diagnostics is further expanded through image recognition tools, including detection \cite{lin2020two} and classification \cite{dai2023addressing}, which are increasingly being used in conjunction with clinical textual reports to provide a more holistic understanding of medical conditions. Initiatives such as Quilt-1m \cite{ikezogwo2024quilt}, and pre-training projects like BiomedCLIP \cite{zhang2023large} and MedClip \cite{wang2022medclip}, as well as efforts in 3D image description by GTGM \cite{chen2023generative}, T3D \cite{liu2023t3d}, and 3d-MIR \cite{abacha20233d}, reflect the ongoing efforts to connect medical images with their textual counterparts. Despite these advancements, the focus on 2D image-text pairs and the challenges with model-generated descriptions highlight the need for a comprehensive benchmark that integrates 3D medical imaging with textual diagnostics, a significant void in the current landscape \cite{wantlin2023benchmd}.

Building on these efforts, we introduce {BIMCV-R}, a new endeavor that directly addresses the need for a unified benchmark in 3D medical imaging diagnostics. Utilizing the comprehensive medical data repository BIMCV \cite{vaya2020bimcv} and in collaboration with clinicians, we curate \textbf{the first publicly accessible dataset} that features 8,069 3D medical image-report pairs, covering 96 disease types. Conscious of privacy concerns, we anonymize radiological reports and translate the original Spanish dataset into English using GPT-4, with meticulous human proofreading to ensure accuracy and reliability.
Furthermore, we develop MedFinder, a dual-stream network architecture that leverages the advanced capabilities of the large language model BiomedCLIP \cite{zhang2023large} to establish a bridge between medical images and reports through text-image retrieval tasks. This initiative lays the foundation for benchmarks in text-image and keyword retrieval, significantly simplifying the process for physicians and clinicians to search for and reference similar cases, thereby enhancing diagnostic precision and efficiency.
% Furthermore, we established a bridge between medical images and reports via text-image retrieval tasks, employing the advanced capabilities of the large language model BiomedCLIP \cite{zhang2023large} %and the SOTA 3D medical image model GTGM
% . This initiative lays the groundwork for benchmarks in text-image and keyword retrieval, significantly streamlining the process for physicians and radiologists to search for and reference similar cases, thereby boosting diagnostic precision and efficiency.

Our contributions are manifold and significant: 
\begin{enumerate}
    \item We curate the {first publicly accessible} English 3D text-image CT dataset {BIMCV-R}, inclusive of authentic radiological reports and detailed disease-type diagnoses.
    \item We introduce MedFinder, an exhaustive suite of medical retrieval schemes, including innovative approaches for text-image, image-text, and keyword-image retrieval—a pioneering effort on a real-world dataset.
    \item By harnessing the power of pre-trained large language models, we showcase their untapped potential in enhancing 3D medical image retrieval, thereby filling a critical void in the field and setting a new standard for future research and application in medical image analysis and retrieval.
\end{enumerate}

\section{Dataset}
This paper presents the BIMCV-R dataset, a substantial resource meticulously crafted for 3D medical multimodal retrieval. This dataset is an extension of the BIMCV dataset \cite{vaya2020bimcv}, encompassing pristine CT scan images, detailed radiological reports, and comprehensive DICOM metadata. 
% We have secured permission from the BIMCV team and are committed to releasing the dataset to the public upon the publication of this paper.
\begin{figure}[t]
    \centering
    \includegraphics[width=\linewidth]{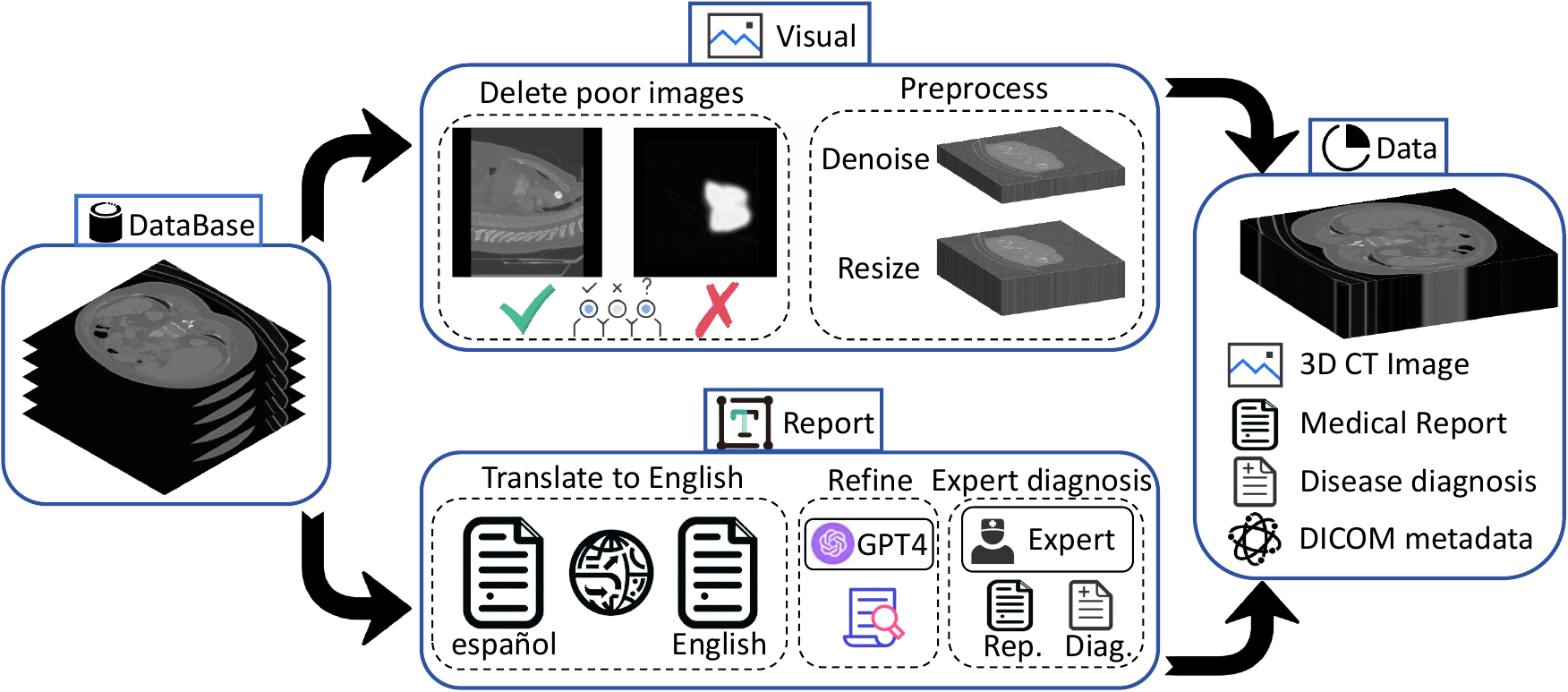}
    \caption{Construction of the BIMCV-R dataset. Utilizing the BIMCV dataset, we enhanced image quality through selective filtering, advanced denoising, and size standardization. For textual data, we translated radiological reports into English and refined them with GPT-4, ensuring consistency. Expert reviews and diagnoses further ensured data reliability and accuracy.}
    % \vspace{-0.2cm}
    \label{fig:main1}
\end{figure}
\subsubsection{Data Acquisition and Processing.}
The acquisition process of BIMCV-R, as illustrated in Figure \ref{fig:main1}, commenced with the initial phase of our dataset processing where we eliminated image instances with pixel missing values exceeding 30\%, and discarded CT scan samples with any dimension (width, height, or depth) less than 96. Subsequently, CT images of superior imaging quality were manually selected, and images lacking corresponding medical descriptions were removed. Regarding the textual content, descriptions shorter than five words were omitted, and personal information within the text descriptions, such as names and addresses, was anonymized. Following this, all textual descriptions were translated into English using GPT-4 and underwent a manual verification process, culminating in a dataset comprising 8,069 paired samples with more than 2M slices. More importantly, we engaged over 20 medical professionals to diagnose 1,475 of these samples, identifying 96 different diseases, including tumors, infectious diseases, cardiovascular diseases, and respiratory conditions. These diagnoses facilitated the development of an extensive keyword library. Ultimately, we constructed a dataset exceeding 700GB, encompassing original CT scan images, radiological reports, and DICOM metadata, offering a comprehensive resource for medical research and application development in the field of medical imaging.

\subsubsection{Data Statistics Analysis.}
\begin{figure}[t]
    \begin{minipage}[b]{0.4\linewidth}
        \centering
        \captionof{table}{Summary of Image and Report Statistics.}
        % \vspace{-0.4cm}
        \fontsize{6.5}{8}\selectfont
        \centering
        \renewcommand\tabcolsep{0.5pt}
        \begin{tabular}{cccc}
            \toprule[1.2pt]
            \textbf{Statistic} & \textbf{Average} & \textbf{Median} & \textbf{Range} \\ \hline
            Image Width & 529 & 512 & 514 $\sim$ 710 \\
            Image Height & 528 & 512 & 520 $\sim$ 672 \\
            Number of Slices & 279 & 224 & 101 $\sim$ 670 \\
            Length of Report & 104 & 97 & 7 $\sim$ 260 \\ \bottomrule[1.2pt]
        \end{tabular}
        \label{tab:stats}
    \end{minipage}%
    \hfill
    \begin{minipage}[t]{0.55\linewidth}
        \centering
        \includegraphics[width=\linewidth]{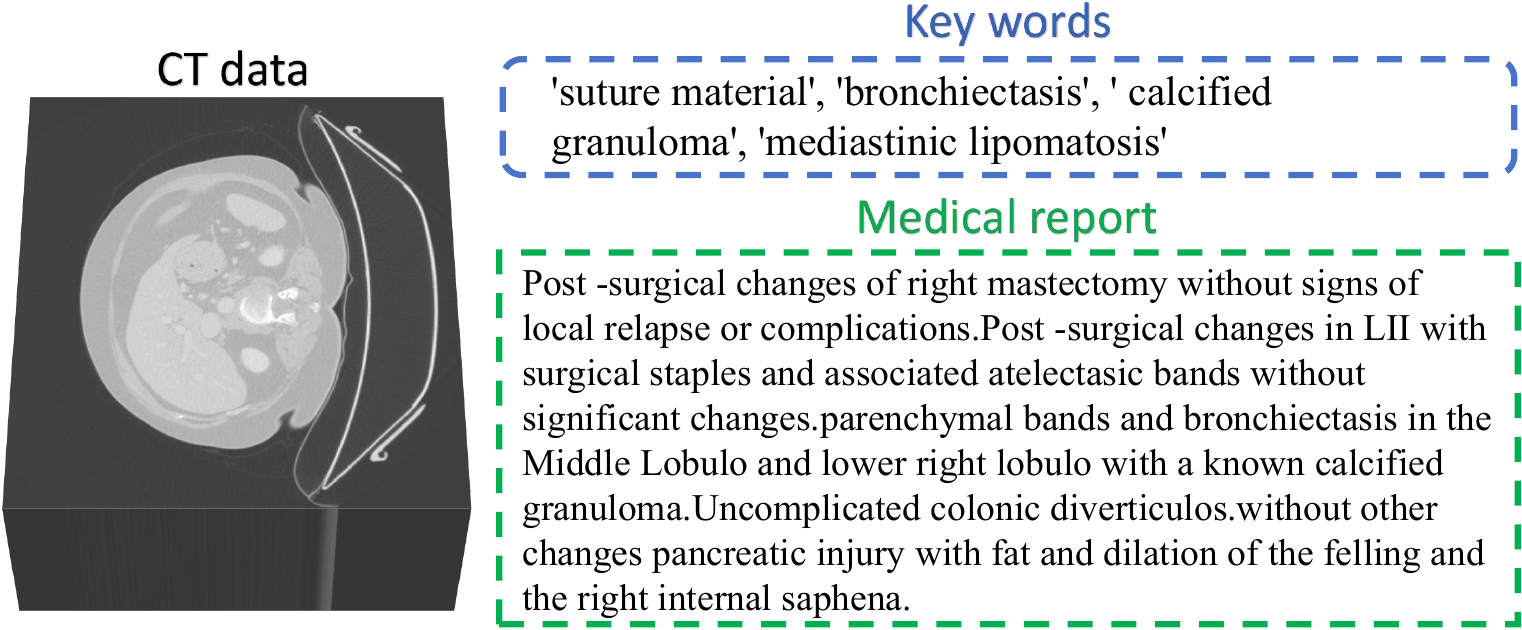}
        \vspace{-0.6cm}
        \caption{Sample data of BIMCV-R.}
        \label{fig:sample}
    \end{minipage}
    \vspace{-0.3cm}
\end{figure}

\begin{figure}[t]
    \centering
    \begin{minipage}{0.4\textwidth}
        \centering
        \includegraphics[width=\linewidth]{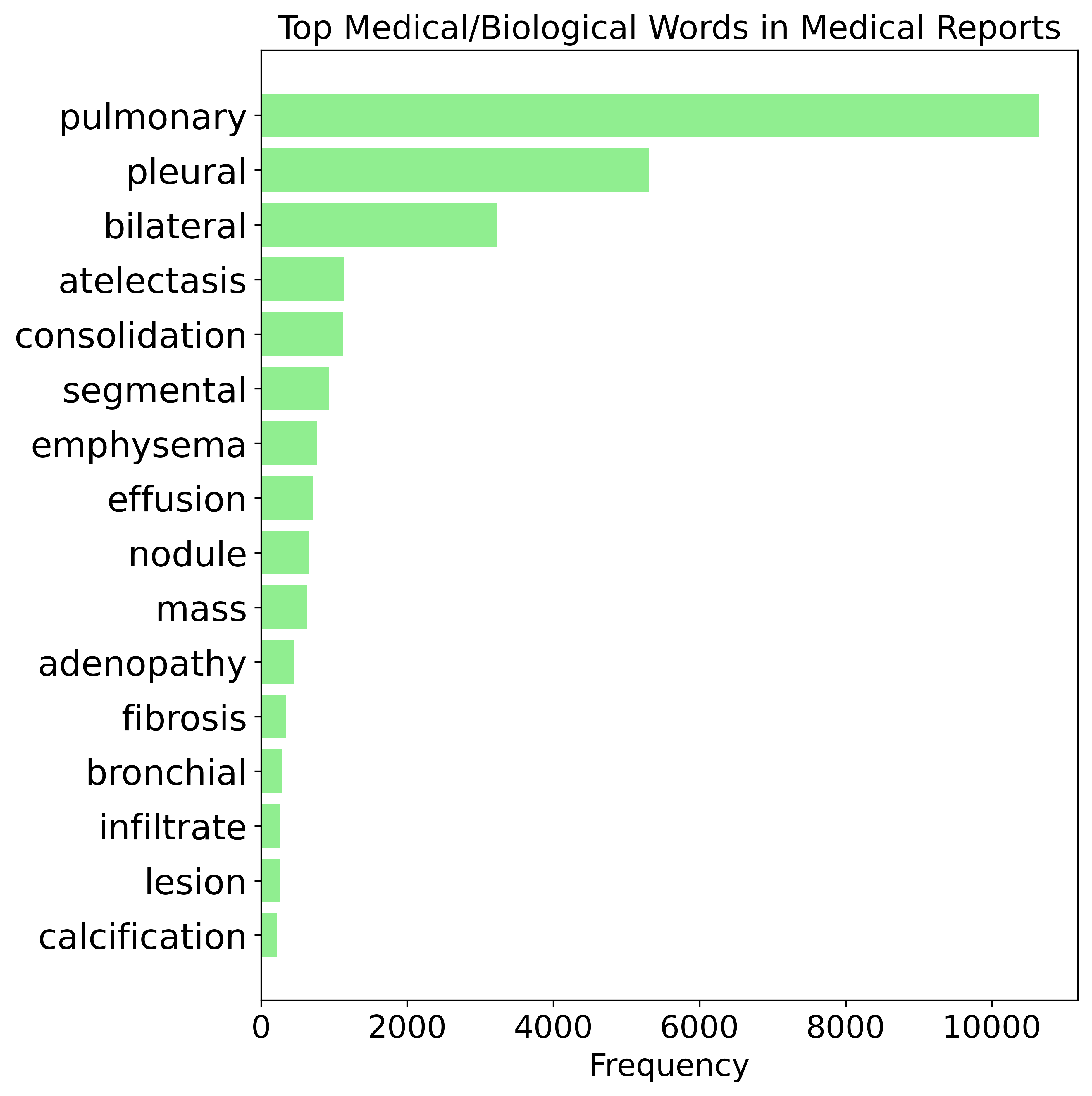}
        % \caption{Your first caption here}
        \label{fig:your-first-label}
    \end{minipage}\hfill
    \begin{minipage}{0.6\textwidth}
        \centering
        \includegraphics[width=\linewidth]{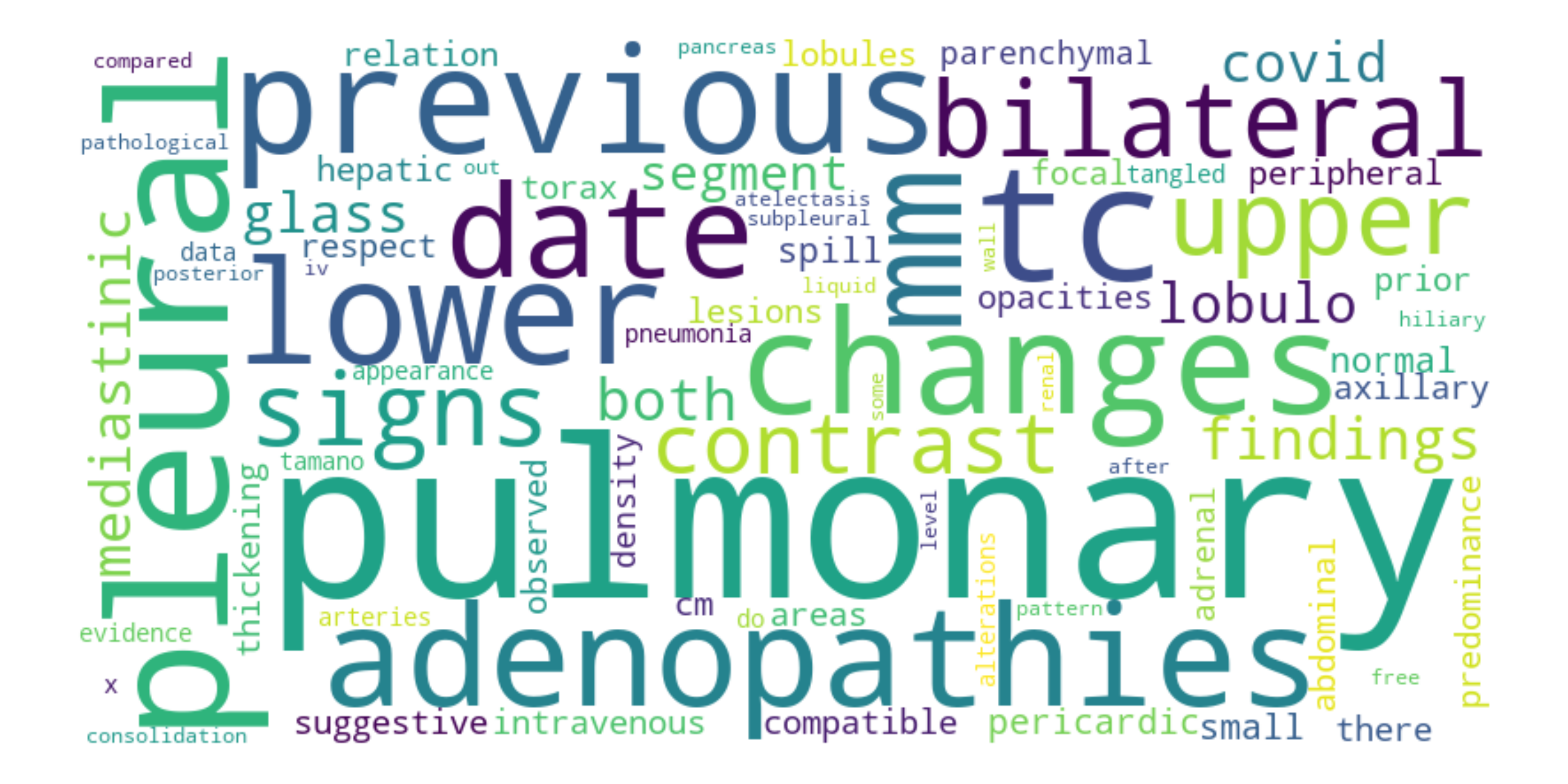}
        % \caption{Your second caption here}
        \label{fig:your-second-label}
    \end{minipage}
    \vspace{-0.5cm}
    \caption{\textbf{Left:} Word Frequency Analysis.
    \textbf{Right:} World Cloud Analysis.}
    % \vspace{-0.5cm}
    \label{fig:frequency}
\end{figure}

The BIMCV-R dataset is distinguished by its inclusion of high-resolution 3D medical images paired with corresponding radiological reports, providing a rich resource for training deep learning models capable of understanding and processing 3D medical imagery. Through this approach, models can learn the correlation between visual features extracted from images and the linguistic descriptions found in radiological reports. We have compiled basic statistics of the dataset as shown in Table \ref{tab:stats}, and we present sample data as illustrated in Figure \ref{fig:sample}. Furthermore, based on diagnoses from medical experts, we have conducted a keyword frequency analysis of the radiological reports, with the results depicted in Figure \ref{fig:frequency}. 
% This methodology allows for an enhanced understanding of the relationship between imaging findings and their clinical interpretations, facilitating the development of more accurate and insightful medical diagnostic models.

\section{Methodology}
% \vspace{-0.2cm}
\subsubsection{Overview.}
\begin{figure}[t]
    \centering
    \includegraphics[width=\linewidth]{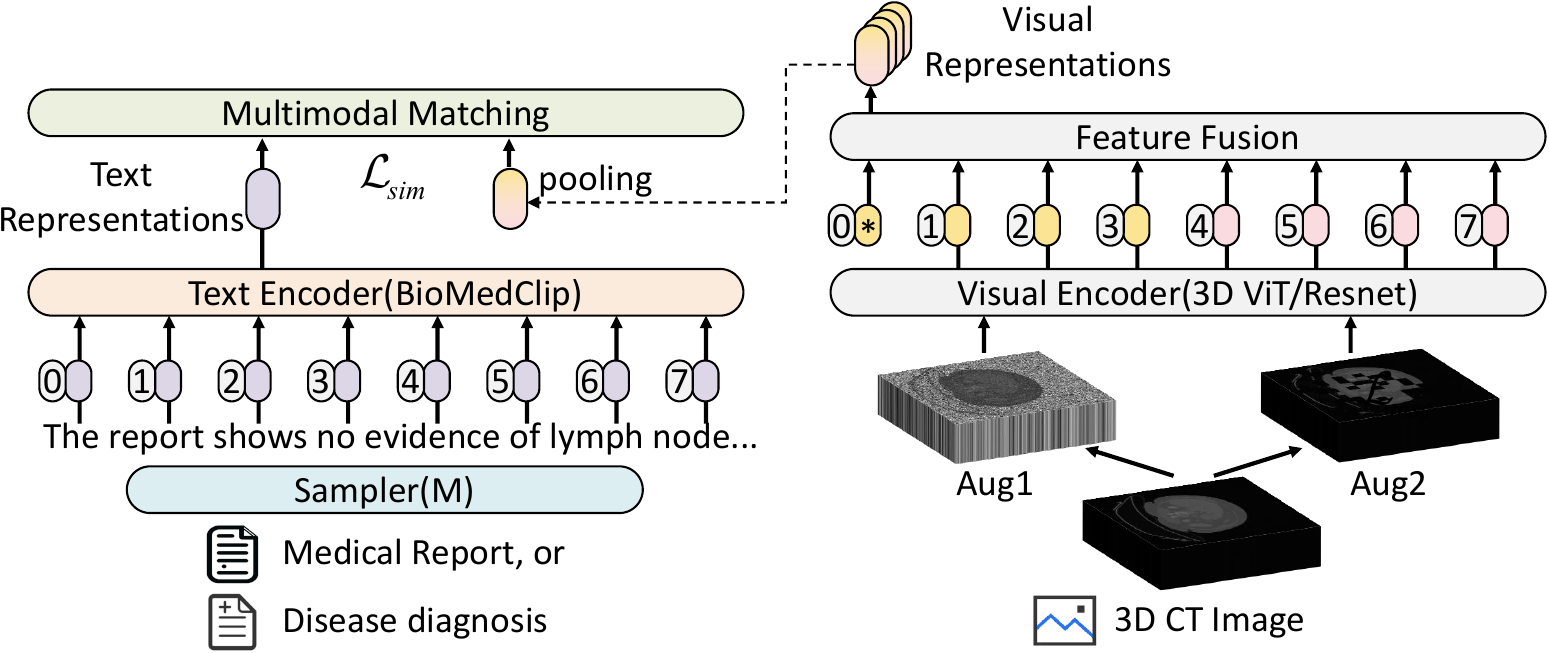}
    \caption{An overview of our method, divided into textual feature extraction, visual feature extraction, and similarity matching.}
    % \vspace{-0.4cm}
    \label{fig:main}
\end{figure}

The comprehensive workflow of MedFinder is depicted in Figure \ref{fig:main}. Initially, we sample the lengthy textual description \(T\) for manageable processing. For the text \(T\), we employ a sampler \(S\) to randomly select \(M\) words, where \(M\) is typically set to 64, resulting in a sampled textual representation \(T' = S(T, M)\). Subsequently, for a 3D medical image \(I\), we apply a 3D image encoder \(F_{\text{3D}}\) to extract its feature representation \(Z = F_{\text{3D}}(I)\).
To enhance the model's discriminative capability for medical image features, we introduce the concept of view consistency. Specifically, we apply two different data augmentation techniques \(A_1\) and \(A_2\) to the original 3D medical image \(I\), generating two augmented views \(I_1\) and \(I_2\). These views are processed through the 3D image encoder to obtain feature representations \(Z_1 = F_{\text{3D}}(I_1)\) and \(Z_2 = F_{\text{3D}}(I_2)\). A view consistency loss \(L_{\text{cons}}\) ensures the consistency between these two feature representations, aiding the model in learning more robust image feature representations.
Following feature extraction, we employ a feature discrimination loss \(L_{\text{dis}}\) to further refine the model, encouraging it to learn to distinguish between different medical image features. Finally, we integrate textual and image features, using a similarity metric \(S\) to compute the similarity score \(s(T', Z)\) between the text and image.

\subsubsection{Textual Feature Extracting.}
Consider a medical text $T \in \mathbb{R}^{N \times M}$, where $N$ represents the maximum length of the text, and $M$ denotes the size of the vocabulary. Our objective is to extract features from this text to facilitate subsequent image retrieval tasks.

Initially, we define a sampling function $S: \mathbb{R}^{N \times M} \times \mathbb{N} \rightarrow \mathbb{R}^{L \times M}$, which selects a continuous text segment of length $L$ (here, $L = 100$) from the input text $T$. This sampling process can be represented as $T' = S(T, L)$,
where $T'$ is the sampled text segment, a matrix of size $L \times M$, representing 100 selected words and their corresponding vocabulary indices from the original text.

Subsequently, we employ a pre-trained text encoder $F_{\text{biomedClip}}: \mathbb{R}^{L \times M} \rightarrow \mathbb{R}^{D}$ to extract features from the text segment. The encoder $F_{\text{biomedClip}}$ is a deep neural network that maps the text segment $T'$ to a $D$-dimensional feature space. The feature extraction process can be represented as
$W = F_{\text{biomedClip}}(T'),
$
where $W$ is the feature vector of the text segment $T'$, a $D$-dimensional vector containing semantic information of the text segment. We freeze the text encoder during training.

% During the training process, we keep the parameters of the text encoder $F_{\text{biomedClip}}$ fixed, implying no further training is conducted on it. This approach is aimed at leveraging the pre-trained model's knowledge in the biomedical domain while reducing the complexity of training. We can freeze the model parameters as follows:
% \[
% \theta_{F_{\text{biomedClip}}} = \text{const}
% \]
% where $\theta_{F_{\text{biomedClip}}}$ represents the parameters of the text encoder.

\subsubsection{Visual Feature Extracting.}
Consider a 3D medical image block \(I \in \mathbb{R}^{H \times W \times D}\), where \(H\), \(W\), and \(D\) represent the dimensions, our goal is to extract features for retrieval tasks.
We use a resizing function \(R: \mathbb{R}^{H \times W \times D} \rightarrow \mathbb{R}^{H' \times W' \times D'}\), to resize \(I\) to a standard size \(H' \times W' \times D'\).

Subsequently, we apply two data augmentation techniques \(A_1\) and \(A_2\), such as noise addition, rotation, and cutmix, to improve generalization. The augmented images are 
\begin{equation}
    I_{\text{aug1}} = A_1(I'), \quad I_{\text{aug2}} = A_2(I'),
\end{equation}
where \(I_{\text{aug1}}\) and \(I_{\text{aug2}}\) are the enhanced images.

A visual encoder \(F_{\text{3D}}: \mathbb{R}^{H' \times W' \times D'} \rightarrow \mathbb{R}^{D''}\) extracts features from these images, producing 
\begin{equation}
    Z_{\text{aug1}} = F_{\text{3D}}(I_{\text{aug1}}), \quad Z_{\text{aug2}} = F_{\text{3D}}(I_{\text{aug2}}),
\end{equation}
with \(Z_{\text{aug1}}\) and \(Z_{\text{aug2}}\) as the feature vectors.

To capture class-discriminative features, we obtain the CLS feature vector \(Z_{\text{cls}}\) from the encoder's output using
\begin{equation}
    Z_{\text{cls}} = \text{CLS}(F_{\text{3D}}(I')),
\end{equation}
where \(\text{CLS}\) extracts class-discriminative features.

During training, we ensure semantic consistency of augmented features by computing the Mean Squared Error (MSE) loss between them.
 The MSE loss is articulated as
\begin{equation}
L_{\text{mse}} = \frac{1}{2} \| Z_{\text{aug1}} - Z_{\text{aug2}} \|^2_2,
\end{equation}
where \(\| \cdot \|_2\) signifies the L2 norm.

After obtaining feature vectors \(Z_{\text{aug1}}\) and \(Z_{\text{aug2}}\). We employ a cross-attention mechanism to merge these features, allowing the model to link the vectors for enhanced feature representation.

We define a cross-attention function \(\text{CrossAttn}: \mathbb{R}^{D''} \times \mathbb{R}^{D''} \rightarrow \mathbb{R}^{D''}\), which inputs two feature vectors and outputs a fused feature vector as \(Z_{\text{fusion}} = \text{CrossAttn}(Z_{\text{aug1}}, Z_{\text{aug2}})\).

The calculation of cross-attention is detailed as follows:
% \begin{equation}
% \text{Attention}(Z_{\text{aug1}}, Z_{\text{aug2}}) = \text{softmax}\left(\frac{Z_{\text{aug1}}^T Z_{\text{aug2}}}{\sqrt{D''}\|Z_{\text{aug1}}\|\|Z_{\text{aug2}}\|}\right), \\
% Z_{\text{fusion}} = Z_{\text{aug1}} \odot \text{Attention}(Z_{\text{aug1}}, Z_{\text{aug2}}),
% \end{equation}
% \begin{align}
% \text{Attention}(Z_{\text{aug1}}, Z_{\text{aug2}}) &= \text{softmax}\left(\frac{Z_{\text{aug1}}^T Z_{\text{aug2}}}{\sqrt{D''}\|Z_{\text{aug1}}\|\|Z_{\text{aug2}}\|}\right), \\
% Z_{\text{fusion}} &= Z_{\text{aug1}} \odot \text{Attention}(Z_{\text{aug1}}, Z_{\text{aug2}}),
% \end{align}
\begin{gather}
\text{Attention}(Z_{\text{aug1}}, Z_{\text{aug2}}) = \text{softmax}\left(\frac{Z_{\text{aug1}}^T Z_{\text{aug2}}}{\sqrt{D''}\|Z_{\text{aug1}}\|\|Z_{\text{aug2}}\|}\right), \\
Z_{\text{fusion}} = Z_{\text{aug1}} \odot \text{Attention}(Z_{\text{aug1}}, Z_{\text{aug2}}),
\end{gather}
where \(\odot\) represents the element-wise multiplication (Hadamard product), and \(\text{softmax}\) is a normalization function ensuring the sum of attention weights equals 1. Thus, a feature vector \(Z_{\text{fusion}}\) that merges information from two views is derived.

\subsubsection{Similarity Matching.}
Upon obtaining the visual features \(Z_{\text{fusion}}\) and text features \(W\), we proceed with the following steps for retrieval.

Initially, we perform a pooling operation on the visual features \(Z_{\text{fusion}}\) to obtain a fixed-length feature vector, expressed as
$
Z_{\text{pooled}} = \frac{1}{D''} \sum_{i=1}^{D''} Z_{\text{fusion},i},
$
where \(Z_{\text{pooled}}\) is the pooled visual feature vector, and \(D''\) represents the dimensionality of the visual features.

Next, the pooled visual features \(Z_{\text{pooled}}\) are matched with the text features \(W\) to compute their similarity score. The similarity score is calculated using cosine similarity, denoted as
$
\text{Score}(W, Z_{\text{pooled}}) = \frac{W^T Z_{\text{pooled}}}{\|W\| \|Z_{\text{pooled}}\|},
$
where \(\text{Score}\) is the similarity score between the text features \(W\) and the pooled visual features \(Z_{\text{pooled}}\).

During training, we have a set of positive pairs \((W_i, Z_{\text{pooled},i})\) and negative pairs \((W_j, Z_{\text{pooled},j})\), the loss function is represented as:
\begin{equation}
L_{sim} = -\sum_{i,j} \log \frac{\exp(\text{Score}(W_i, Z_{\text{pooled},i}))}{\sum_{k} \exp(\text{Score}(W_k, Z_{\text{pooled},i}))},
\end{equation}
where \(L_{sim}\) is the loss function, encouraging the model to increase the similarity score of positive pairs while decreasing that of negative pairs \cite{li2023lvit}.

The total loss can be written as
\begin{equation}
L_{\text{total}} = L_{\text{mse}} + \alpha L_{sim},
\end{equation}
where \(L_{\text{total}}\) is the total loss, combining MSE loss (\(L_{\text{mse}}\)) and matching loss (\(L_{sim}\)), weighted by \(\alpha\).
\section{Experiments and Results}
% \subsection{Dataset and Implementation Details}
\subsubsection{Data Splitting and Metrics.}
In this study, we configured the BIMCV-R dataset into training, validation, and test sets, accounting for 70\%, 10\%, and 20\% of the total data, respectively. To thoroughly evaluate the performance of our multimodal retrieval system, we selected Recall@K (R@K), Median Rank (MdR), and Mean Rank (MnR) as our primary evaluation metrics. These metrics provide a multidimensional reflection of the model's effectiveness in retrieval tasks. Furthermore, to assess the performance of keyword-based image retrieval, we incorporated the Precision@K (P@K) metric, which precisely evaluates the proportion of correct items in the returned results. In our performance comparison experiments, we chose the CLIP4Clip \cite{luo2022clip4clip} model from the video retrieval domain and the 3D-MIR \cite{abacha20233d} model from the medical multimodal retrieval domain as benchmarks to validate the effectiveness of our approach.

\subsubsection{Results.}
We conducted experiments on both multimodal retrieval and keyword-based retrieval, where the CLIP4Clip and 3D-MIR models processed 3D medical volumes using frame-by-frame and average input methods, respectively, as described in the original studies. Our approach provided experimental results using ResNet-50 and ViT-base as backbones. The results of the multimodal retrieval experiments are shown in Table \ref{tab:multimodalR}. These results indicate that our method outperforms the baselines. Specifically, CLIP4Clip's performance was compromised due to the use of CLIP weights in the text processing component, which has a significant gap with medical descriptions, leading to inferior results. Despite significant improvements in our method, the lengthy nature of medical diagnostics and the high similarity among input images mean there is still a noticeable gap compared to image-text retrieval and text-video retrieval tasks.
% \usepackage{multirow}
\begin{table}[t]
 \fontsize{7.5}{8.5}\selectfont
        \centering
        \renewcommand\tabcolsep{0.5pt}
        \caption{Results of multimodal retrieval, the best results are highlighted in \textbf{bold}.}
\begin{tabular}{l|ccccc|ccccc}
\toprule[1.2pt]
\multirow{2}{*}{Methods} & \multicolumn{5}{c}{Text to Image} & \multicolumn{5}{c}{Image to Text} \\
                         & R@1 $\uparrow$ & R@5 $\uparrow$ & R@10 $\uparrow$  & MdR $\downarrow$  & MnR $\downarrow$   & R@1 $\uparrow$  & R@5 $\uparrow$ & R@10 $\uparrow$ & MdR $\downarrow$   & MnR $\downarrow$   \\ \midrule
CLIP4clip \cite{luo2022clip4clip}                & 0.3 & 1.5 & 2.2   & 717.0 & 735.9 & 0.3  & 0.8 & 1.5  & 722.0 & 738.7 \\
3D-MIR \cite{abacha20233d}                   & 1.1 & 4.7 & 10.3  & 121.1 & 152.3 & 1.2  & 4.0 & 8.8  & 134.9 & 162.4 \\
MedFinder (Resnet-50)         & \textbf{2.8} & 8.7 & 20.3 & 68.9  & \textbf{81.3}  & \textbf{2.9}  & 8.8 & 19.7 & 71.2  & 80.7  \\
MedFinder (ViT-base)          & 2.7 & \textbf{8.9} & \textbf{21.4}  & \textbf{75.4}  & 80.1  & 2.7  & \textbf{9.0} & \textbf{20.3} & \textbf{72.3}  & \textbf{81.9}  \\ \bottomrule[1.2pt]
\end{tabular}

% \vspace{-0.5cm}
\label{tab:multimodalR}
\end{table}
\begin{table}[t]
 \fontsize{8}{9.6}\selectfont
        \centering
        \renewcommand\tabcolsep{2.5pt}
\caption{Results of keyword retrieval, the best results are highlighted in \textbf{bold}.}        
\begin{tabular}{l|ccc|ccc|ccc}
\toprule[1.2pt]
\multirow{2}{*}{Methods} & \multicolumn{3}{c}{atelectasis} & \multicolumn{3}{c}{consolidation} & \multicolumn{3}{c}{adenopathy} \\
                         & P@20     & P@50     & P@100     & P@20      & P@50      & P@100     & P@20     & P@50     & P@100    \\ \midrule
CLIP4Clip                & 0.20     & 0.24     & 0.21      & 0.15      & 0.14      & 0.18      & 0.20     & 0.18     & 0.19     \\
3D-MIR                   & 0.35     & 0.34     & 0.31      & 0.40      & 0.36      & 0.39      & 0.35     & 0.34     & 0.37     \\
MedFinder (Resnet-50)         & \textbf{0.75}     & \textbf{0.72}     & \textbf{0.69}      & \textbf{0.70}      & \textbf{0.66}      & \textbf{0.62}      & \textbf{0.75}     & \textbf{0.72}     & \textbf{0.69}     \\
MedFinder (ViT-base)          & 0.70     & 0.68     & 0.63      & 0.65      & 0.64      & 0.61      & 0.70     & 0.68     & 0.63     \\ \bottomrule[1.2pt]
\end{tabular}

\label{tab:keywordR}
% \vspace{-0.5cm}
\end{table}

To further assess the performance of our model, we employed a keyword-based retrieval task, which aligns more closely with practical scenarios than the rigid CT image-text pairings. Physicians often rely on specific keywords to search for similar cases. Accordingly, we selected three typical diagnoses—atelectasis, consolidation, and adenopathy—as keywords to retrieve related 3D CT slices. As shown in Table \ref{tab:keywordR}, our approach achieves an accuracy of approximately 70\% in retrieving relevant cases, surpassing other baseline methods. This capability can significantly reduce the workload of physicians in real diagnostic processes. Further experimental details and visualization results will be included in the supplementary materials, with the code to be made available open source upon the publication of the paper.
\subsubsection{Ablation Study.}
Our ablation studies, detailed in Table \ref{tab:ab}, show that the Text Sampler and using pretrained text encoder weights significantly affect performance. This is likely due to the detailed nature of retrieval texts, where text sampling enables the network to learn more comprehensive descriptions. Additionally, a medical-specific text encoder extracts more relevant information, essential for our task.
% \usepackage[normalem]{ulem}
\begin{table}[!t]
 \fontsize{8}{9.6}\selectfont
        \centering
        \renewcommand\tabcolsep{3pt}
\caption{Ablation study in text to image retrieval tasks using Resnet-50 backbone.}
\begin{tabular}{ccc|ccccc}
\toprule[1.2pt]
{Text Sampler} & {$L_{mse}$} & {Text Encoder} & {{R@1}} & {{R@5}} & {{R@10}} & MdR & MnR \\ \midrule
                                    & $\checkmark$                              & Clip                                & 1.2                              & 5.1                              & 11.3 & 120.3  &  148.4                             \\
$\checkmark$                                 &                                  & Clip                                & 1.3                              & 5.4                              & 11.8       & 117.3  & 142.5                       \\
$\checkmark$                                 & $\checkmark$                              & Clip                                & 1.8                              & 6.3                              & 13.9     & 107.9 & 129.6                         \\
\rowcolor[HTML]{EFEFEF}  
$\checkmark$                                 & $\checkmark$                              & BiomedCLIP                          & 2.8                              & 8.7                              & 20.3 &68.9 & 81.3     \\ \bottomrule[1.2pt]                      
\end{tabular}

% \vspace{-0.5cm}
\label{tab:ab}
\end{table}

\section{Conclusion}
% In this study, we introduce the BIMCV-R dataset, aimed at establishing a benchmark for 3D medical image-text retrieval. With a carefully curated collection of 8,069 3D CT slices and their corresponding medical reports, we offer a valuable resource to researchers. Our MedFinder demonstrates that effective information retrieval can be achieved in multimodal and keyword retrieval tasks by integrating advanced language models with image processing technologies. While our findings provide new insights into this field, many challenges remain to be addressed. We hope the BIMCV-R dataset will inspire further research to advance the development of 3D medical image analysis technologies.
In this study, we introduce the BIMCV-R dataset, aimed at establishing a benchmark for 3D medical image-text retrieval. With a carefully curated collection of 8,069 3D CT volumes and their corresponding radiological reports, we offer a valuable resource to researchers. Our MedFinder demonstrates that effective information retrieval can be achieved in multimodal and keyword retrieval tasks by integrating advanced language models with image processing technologies. While our exploration is in its initial stages, it opens up a novel direction for the field, underscoring the potential for fruitful advancements in 3D medical image analysis technologies. We hope the BIMCV-R dataset will inspire further research to advance the development of 3D medical image analysis technologies.

\section*{Disclosure of Interests}
The authors have no competing interests to declare that are relevant to the content of this article.
% \section*{Acknowledgement}
% This work was supported in part by the National Natural
% Science Foundation of China under Grant 62021001.

% \clearpage
% \bibliographystyle{plain}
{
\small
% \bibliography{ref}
% \bibliographystyle{ieee}
\bibliographystyle{unsrt}

% \bibliography{ref}
}
\end{document}